\documentclass[10pt,journal,compsoc]{IEEEtran}

\ifCLASSOPTIONcompsoc
  \usepackage[nocompress]{cite}
\else
  \usepackage{cite}
\fi

\usepackage{cite}
\usepackage{enumitem}
\usepackage{ragged2e} 
\ifCLASSINFOpdf
\else
\fi

\usepackage{ragged2e}
\usepackage{graphicx}
\usepackage{booktabs}
\usepackage{makecell} 

\begin{document}
%
\title{OPEN: A Benchmark Dataset and Baseline for Older Adult Patient Engagement Recognition in Virtual Rehabilitation Learning Environments}

\author{Ali~Abedi,        
        Sadaf~Safa,
        Tracey~J.F.~Colella\textsuperscript{\dag},
        Shehroz~S.~Khan\textsuperscript{\dag}
\IEEEcompsocitemizethanks{\IEEEcompsocthanksitem Ali Abedi, Sadaf Safa, Tracey J.F. Colella, and Shehroz S. Khan were with the KITE Research Institute, Toronto Rehabilitation Institute, University Health Network, Toronto, Canada.\protect

†Shehroz S. Khan and Tracey J.F. Colella are senior authors of this work.
E-mail: ali.abedi@uhn.ca
}
\thanks{Manuscript received July 23, 2025; revised August 31, 2025.}}

%
%

\markboth{IEEE Transactions on Learning Technologies,~Vol.~18, No.~18, July~2025}%
{Shell \MakeLowercase{\textit{et al.}}: Bare Advanced Demo of IEEEtran.cls for IEEE Computer Society Journals}

\IEEEtitleabstractindextext{%
\begin{abstract}
\justifying
Engagement in virtual learning is critical to participant satisfaction, performance, and adherence, particularly in domains such as online education and virtual rehabilitation, where interactive communication is essential for success. However, accurately measuring engagement in virtual group settings remains a significant challenge. There is growing interest in leveraging artificial intelligence for large-scale, real-world, and automated engagement recognition. Although engagement has been widely studied among younger populations in academic contexts, research methods and datasets focused on older adults in virtual and telehealth learning environments remain limited. Existing approaches often overlook the contextual relevance of learning materials and the longitudinal dynamics of engagement across sessions. This paper presents OPEN (Older adult Patient ENgagement), a novel dataset created to support the development of artificial intelligence models for engagement recognition. The dataset was collected from eleven older adults participating in virtual group learning sessions over six weeks as part of their cardiac rehabilitation programs, yielding over 35 hours of data, which represents the largest dataset of its kind. While raw video is withheld to protect privacy, the publicly released data include facial, hand, and body joint landmarks, as well as behavioral and affective features extracted from video. Observational annotations comprise second-level binary engagement states, emotional and behavioral labels, and context-type details, such as whether the instructor addressed the group or an individual. Multiple versions of the dataset were generated using sample lengths of 5, 10, and 30 seconds, as well as variable-length segments. To demonstrate its utility, various machine learning and deep learning models were trained on the annotated data, achieving engagement recognition accuracy as high as 81\%. OPEN provides a scalable foundation not only for advancing personalized, artificial intelligence-driven engagement recognition in aging populations but also for contributing to broader engagement recognition research.
\end{abstract}

}

\maketitle

\IEEEdisplaynontitleabstractindextext

%
\IEEEpeerreviewmaketitle

\ifCLASSOPTIONcompsoc
\IEEEraisesectionheading{\section{Introduction}\label{sec:introduction}}
\else
\section{Introduction}
\label{sec:introduction}
\fi

%
%
%
%
\IEEEPARstart{R}{ehabilitation} aims to improve recovery, reduce disability, and optimize health outcomes through exercise, education, and counseling \cite{WHORehabilitation}. However, traditional in-person rehabilitation faces several barriers to attendance, including transportation difficulties, scheduling conflicts, financial constraints, and healthcare staff shortages, all of which contribute to high dropout rates \cite{shanmugasegaram2012psychometric}. Virtual rehabilitation, a subset of telehealth, refers to the remote delivery of rehabilitation services through digital technologies. It provides home-based educational and exercise sessions as an alternative to in-person programs \cite{nabutovsky2024home, boukhennoufa2022wearable, naeemabadi2020telerehabilitation, rahman2022ai}. The educational components typically involve clinician-led group sessions in which patients receive guidance on managing chronic conditions, adopting healthy behaviors, and developing self-management skills for sustainable care. Research indicates that virtual rehabilitation programs can achieve outcomes comparable to those of in-person programs and help overcome access-related barriers. Furthermore, artificial intelligence (AI) is increasingly integrated into these platforms to enhance patient assessment, monitor activity, and support health outcome prediction \cite{abedi2024artificial}.

Patient engagement in rehabilitation is a dynamic process that evolves over time \cite{bright2015conceptual}. It has been defined as "an increased motivation, attention, and active participation in rehabilitation, grounded in and supported by the interaction and relationship between the patient and clinician" \cite{danzl2012facilitating}. Engagement is critical to the success of rehabilitation programs, as higher engagement is associated with better adherence and reduced dropout rates \cite{bright2015conceptual}. Continuously monitoring engagement throughout the program and implementing timely interventions to support it can significantly enhance participation and lead to improved health outcomes \cite{bright2015conceptual}.

In a comprehensive literature review on the conceptualization and definition of patient engagement in virtual telehealth sessions, Liu et al. \cite{liu2024unpacking} identified three core components of engagement: affective, behavioral, and cognitive. This framework aligns with earlier work in educational psychology by Fredricks et al. \cite{fredricks2004school}, who described engagement as a proxy for involvement and interaction in the student learning context. Affective engagement refers to emotional responses such as excitement and interest; behavioral engagement involves observable actions such as attendance, participation, and staying on task; and cognitive engagement reflects an individual’s investment in learning and willingness to exert effort and embrace challenges. Building on this foundation, Sinatra et al. \cite{sinatra2015challenges} introduced the concept of grain size, which defines the level at which engagement is conceptualized and measured. Grain size ranges from macro-level (e.g., group engagement) to micro-level (e.g., momentary individual engagement in a specific task). Micro-level engagement can be assessed using physiological signals such as blink rate, head pose, and heart rate, offering fine-grained insight into a person’s immediate involvement \cite{sinatra2015challenges}. Engagement can also be viewed along a continuum of analysis levels: context-oriented, person-in-context, and person-oriented. Macro-level engagement corresponds to context-oriented analysis, micro-level to person-oriented analysis, and person-in-context lies between the two. Person-in-context engagement captures how individuals interact with specific environmental contexts, such as reading a web page or participating in an online classroom. Complementing these conceptualizations, Salam et al. \cite{salam2023automatic} reviewed context-aware engagement inference across domains, identifying context-aware computational modeling and the temporal dynamics of engagement as key areas for investigation.

There is growing interest in using AI and affective computing to recognize engagement levels in naturalistic settings and at scale \cite{booth2023engagement,dewan2019engagement,salam2023automatic,khan2022inconsistencies,mandia2024automatic}. Current approaches often rely on supervised or semi-supervised machine learning and deep learning methods, which require annotated ground-truth datasets for model development. However, most existing methods and datasets have focused on engagement recognition among young, healthy student populations \cite{dewan2019engagement,salam2023automatic,khan2022inconsistencies,mandia2024automatic}. In contrast, progress in creating datasets and developing AI models tailored to older adult patient engagement has been limited \cite{noceti2024predicting}.

Existing engagement datasets, along with the recognition algorithms developed using them, incorporate a variety of data modalities, including video, head pose, eye gaze, facial expressions, speech, heart rate, and electrocardiogram signals \cite{booth2023engagement,dewan2019engagement,salam2023automatic,khan2022inconsistencies,mandia2024automatic}. The affective, behavioral, and cognitive components of engagement are expressed differently across these modalities depending on factors such as age \cite{guo2013facial}, gender \cite{abbruzzese2019age}, ethnicity, and physical \cite{alonso2014selective} or mental health status \cite{zhou2023developing}. As a result, engagement recognition algorithms trained on data from young, healthy student populations may not generalize well to older adults or patient populations.

To support the development of engagement recognition algorithms that generalize to older adults and patients in virtual learning environments, the primary contribution of this paper is the introduction of OPEN (Older adult Patient ENgagement), a novel public dataset with the following distinctive features:

\begin{itemize}
    \item Targets the older adult patient population and constitutes the largest publicly available dataset for AI-driven engagement recognition.
    \item Includes annotated engagement states along with affective and behavioral components, captured by trained observers, and provides contextual information about the learning sessions.
    \item Offers multiple versions of the dataset with fixed-length (e.g., 10- and 30-second) and variable-length samples, incorporating both sequential engagement data within sessions and longitudinal data across sessions to support engagement detection and prediction.
    \item Preserves participant privacy by publicly releasing non-identifiable data, including facial, hand, and body joint landmarks, as well as derived behavioral and affective features.
\end{itemize}

As a secondary contribution, a range of advanced machine learning and deep learning models were developed to technically validate the OPEN dataset across multiple experimental settings, achieving high engagement recognition performance and confirming the dataset’s efficacy and suitability for AI model development. In addition, and for the first time in the engagement recognition literature \cite{karimah2022automatic,abedi2024engagement,vedernikov2024tcct,engagenet}, two distinct engagement inference tasks, namely \textit{engagement detection} and \textit{engagement prediction}, were systematically investigated.

The structure of this paper is as follows. Section~\ref{sec:related_work} reviews related work on existing engagement recognition datasets and methods, highlighting the distinctions between the proposed dataset and prior efforts. Section~\ref{sec:dataset} details the dataset collection and annotation procedures, along with its key characteristics. Section~\ref{sec:experiments} outlines the methodology for engagement recognition, describes the experimental settings, and presents the results obtained using the proposed dataset and models. Finally, Section~\ref{sec:conclusion} concludes the paper and discusses directions for future research.

\section{Related Work}
\label{sec:related_work}
This section reviews existing datasets for engagement recognition in virtual learning environments, guided by engagement annotation dimensions defined in prior work \cite{khan2022inconsistencies}. It also provides a brief overview of current engagement recognition methods. Given the limited research focused specifically on older adults and patients in virtual learning settings, the review includes datasets and methods originally developed for student engagement in virtual learning contexts.

\subsection{Dimensions of Engagement Annotation}
\label{sec:engagement_annotation_dimensions}
D’Mello \cite{d2015influence} identified five dimensions of affect annotation, which later served as the basis for Khan et al. \cite{khan2022inconsistencies} to extend the framework by introducing two additional dimensions. This expansion resulted in a comprehensive set of seven engagement annotation dimensions, underscoring the central role of affect as one of the three core components of engagement.
\footnote{Please note the distinction between the three \textit{components of engagement} detailed in section \ref{sec:introduction} and the seven \textit{dimensions of engagement annotation} described in subsection \ref{sec:engagement_annotation_dimensions}.}. 
The seven dimensions are as follows \cite{d2015influence,khan2022inconsistencies}:

\begin{enumerate}
    \item \textit{Sources}: Identifies who performs the annotation, such as human observers or self-reports.
    \item \textit{Data modality}: Specifies the type of input data used for annotation, such as video recordings or screen captures.
    \item \textit{Timing}: Indicates when the annotation is performed, for example, in real time during a session or retrospectively.
    \item \textit{Temporal resolution (timescale)}: Defines the frequency of annotation, such as every second, every 10 seconds, or once per session.
    \item \textit{Level of abstraction}: Distinguishes whether engagement is annotated as a single overall measure or as multiple components (e.g., affective, behavioral, cognitive).
    \item \textit{Combination}: Describes how engagement components are integrated to derive a unified engagement label.
    \item \textit{Quantification}: Specifies how engagement is numerically represented, such as binary (engaged vs. disengaged) or ordinal levels.
\end{enumerate}

\subsection{Engagement Datasets}
\label{sec:engagement_datasets}
In a critical review, Khan et al. \cite{khan2022inconsistencies} examined thirty-one existing engagement datasets, all of which were collected from students in virtual learning environments. Analyzing these datasets through the lens of the seven engagement annotation dimensions revealed considerable inconsistencies in how engagement was defined and annotated. These inconsistencies pose a significant challenge for researchers aiming to develop generalizable AI models for engagement recognition that are both applicable and comparable across datasets. Notably, only a few datasets relied on pre-established, psychometrically validated instruments \cite{aslan2017human,ocumpaugh2015baker} to define and annotate engagement \cite{alyuz2017unobtrusive,alyuz2021annotating,okur2017behavioral}.

Khan et al. \cite{khan2022inconsistencies} identified that the most common \textit{sources} for engagement annotation were observers (in 21 out of 31 datasets) \cite{whitehill2014faces,aslan2014learner,bosch2016detecting,chen2016hybrid,gupta2016daisee,kamath2016crowdsourced,booth2017toward,okur2017behavioral,alyuz2017unobtrusive,zaletelj2017predicting,kaur2018prediction,alkabbany2019measuring,mohamad2019automatic,alyuz2021annotating,bhardwaj2021application,delgado2021student,ma2021hierarchical,gupta2022facial,jeong2022evaluation,verma2022multi,engagenet}, self-reports (in 7 out of 31 datasets) \cite{chen2015video,monkaresi2016automated,de2019engaged,hutt2019time,vanneste2021computer,buono2022assessing,thomas2022automatic}, and a combination of both observers and self-reports (in 3 out of 31 datasets) \cite{bosch2016using,zheng2021estimation,altuwairqi2021new}. In all datasets that involved observers for annotation, the video served as the primary \textit{data modality} observed retrospectively (\textit{timing}) by the annotators. \textit{Temporal resolution (timescale)} varied inconsistently, ranging from 1 second to 30 minutes. Five datasets \cite{bosch2016detecting,okur2017behavioral,alyuz2017unobtrusive,de2019engaged,hutt2019time,alyuz2021annotating} employed adaptive timescales, where annotations were not made at predefined intervals. Instead, new timestamps were generated only when changes in engagement states occurred, resulting in variable-length data samples.

Regarding the \textit{level of abstraction}, fourteen datasets defined and annotated engagement as a simple variable representing engagement level \cite{aslan2014learner,chen2016hybrid,monkaresi2016automated,kamath2016crowdsourced,booth2017toward,zaletelj2017predicting,vanneste2021computer,bhardwaj2021application,zheng2021estimation,ma2021hierarchical,buono2022assessing,jeong2022evaluation,thomas2022automatic,engagenet}. In contrast, others focused on single \cite{chen2015video,gupta2016daisee,bosch2016detecting,okur2017behavioral,alyuz2017unobtrusive,kaur2018prediction,de2019engaged,hutt2019time,delgado2021student,altuwairqi2021new,gupta2022facial} or multi-component aspects of engagement \cite{bosch2016detecting,alkabbany2019measuring,alyuz2021annotating,verma2022multi,whitehill2014faces,mohamad2019automatic}, with five specifically targeting affective and behavioral components \cite{bosch2016detecting,alkabbany2019measuring,mohamad2019automatic,alyuz2021annotating,verma2022multi}. The \textit{combination} dimension is applicable only to those addressing multiple components of engagement. While four datasets did not combine the individual components of engagement \cite{bosch2016detecting,alkabbany2019measuring,alyuz2021annotating,verma2022multi}, two datasets \cite{whitehill2014faces,mohamad2019automatic} employed a set of rules to combine these components and produce an overall engagement level. In terms of \textit{quantification}, engagement levels were defined as ordinal \cite{whitehill2014faces,chen2015video,gupta2016daisee,kamath2016crowdsourced,zaletelj2017predicting,kaur2018prediction,hutt2019time,alkabbany2019measuring,bhardwaj2021application,zheng2021estimation,altuwairqi2021new,thomas2022automatic,engagenet}, interval \cite{booth2017toward,vanneste2021computer,buono2022assessing,gupta2022facial,thomas2022automatic}, and categorical or dichotomous \cite{aslan2014learner,bosch2016detecting,okur2017behavioral,alyuz2017unobtrusive,alyuz2021annotating,delgado2021student,verma2022multi,chen2016hybrid,monkaresi2016automated,bosch2016using,mohamad2019automatic,jeong2022evaluation}, variables.

Except for \cite{monkaresi2016automated}, which included healthy students aged 20 to 60 with an average age of 34, all other thirty engagement in virtual learning datasets reviewed in \cite{khan2022inconsistencies} were collected from young adult healthy students. The virtual learning programs consisted of offline pre-recorded lectures, working with interactive software, reading and writing activities, and only one online lecture with an instructor present \cite{thomas2022automatic}. Of these datasets, eighteen were collected in the wild, while thirteen were collected in a lab setting. All the datasets were collected from students during single, one-time recording sessions. The number of participants in the datasets had a mean of 47.39 with a standard deviation of 40.40, ranging from a minimum of 6 to a maximum of 137.

\subsubsection{Public Engagement Datasets}
\label{sec:public_engagement_datasets}
Only a limited number of the reviewed datasets were publicly available, all of which were collected from young adults in their twenties enrolled at universities in India. Data were collected during single-session, pre-recorded lectures that participants viewed on their computers in various locations. These datasets include the Dataset for Affective States in E-Environments (DAiSEE) \cite{gupta2016daisee}, which contains 9,068 ten-second samples from 112 students, totaling 25.2 hours of video data. The Emotion Recognition in the Wild Engagement Prediction in the Wild (EmotiW-EW) dataset \cite{kaur2018prediction} consists of 195 video samples, each 2 to 3 minutes long, collected from 78 students and comprising 16.5 hours of video. Additionally, EngageNet \cite{engagenet} includes 11,311 ten-second samples from 127 students, totaling 31.4 hours of video data.

\subsubsection{Older Adult Engagement Datasets}
\label{sec:older_adult_engagement_datasets}
Noceti et al. \cite{noceti2024predicting} collected the first engagement dataset focused on older adults. The study involved 12 healthy female participants aged between 77 and 93 years, each taking part in a single 20-minute virtual session from a retirement home in Italy. The dataset comprises a total of 4 hours of data, collected from virtual sessions in which groups of three older adults participated in creative activities under the guidance of a psychologist. The dataset incorporated multiple modalities, including video, audio, accelerometer data, electrodermal activity, and heart rate. Engagement levels were annotated as integers from 1 to 5, representing the lowest to highest levels of engagement. Features extracted from video data included average motion, head pose, Facial Action Units (FAUs), and facial emotions, while audio features consisted of intensity, pitch, and mel-frequency cepstral coefficients. The extracted features, combined with the physiological data, were concatenated and served as input to a fully connected feed-forward neural network for engagement level regression across two temporal settings, 15-second and 2-minute windows. Two significant findings emerged: (1) better performance was achieved with 15-second windows, and (2) the best results were generally obtained using facial and body features from video, with slight improvements when audio data was added. However, the addition of physiological data alongside video and audio features negatively impacted performance. This dataset is not publicly available.

\subsection{Engagement Recognition}
\label{sec:engagement_recognition}
Dewan et al. \cite{dewan2019engagement} categorized engagement recognition techniques into three types based on participant involvement: manual, semi-automatic, and automatic. Manual methods assess engagement using self-report questionnaires that measure factors such as attention, distraction, excitement, and boredom \cite{o2024user}. Semi-automatic methods infer engagement from participants' performance on practice problems and test questions \cite{beck2004using}. In contrast, automatic methods rely on computer vision and machine learning techniques and require no active input from participants. These automatic approaches are valued for being non-intrusive, broadly applicable, and particularly effective in virtual learning environments.

Karimah and Hasegawa \cite{karimah2022automatic} conducted a systematic review of engagement recognition methods and datasets in virtual learning environments, highlighting a predominant focus on the affective and behavioral components of engagement. This emphasis stems from the reliance of most methods on video data alone, without the integration of audio or contextual information, which limits their ability to estimate cognitive engagement. The majority of these approaches use video datasets annotated by external observers \cite{khan2022inconsistencies} to train end-to-end, feature-based, or landmark-based models.

End-to-end techniques use deep neural networks to process raw video frames directly, learning relevant features through architectures such as 3D Convolutional Neural Networks (CNNs) \cite{gupta2016daisee,abedi2021improving,geng2019learning}, Video Transformers \cite{ai2022class}, and combinations of 2D CNNs with Long Short Term Memories (LSTMs) or Temporal Convolutional Networks (TCNs) \cite{gupta2016daisee,abedi2021improving,selim2022students}. Feature-based approaches involve extracting affective, behavioral, or cognitive indicators of engagement from modalities such as video, either using domain-specific knowledge or pre-trained models \cite{tian2023predicting,xie2025msc}. OpenFace \cite{baltrusaitis2018openface} is widely used for extracting FAUs, gaze direction, and head pose \cite{abedi2023affect,abedi2023bag,thomas2018predicting,engagenet}, while facial embeddings from models such as Emotion Face-Alignment Network (EmoFAN) \cite{toisoul2021estimation} and Masked Autoencoder for facial video Representation LearnINg (MARLIN) \cite{cai2023marlin} are also common \cite{abedi2023affect,engagenet}. These features are typically analyzed using models such as Bag-of-Words (BoW), Recurrent Neural Networks (RNNs), TCNs, Transformers, and ensemble models \cite{abedi2023bag,abedi2023detecting,vedernikov2024tcct,tian2023predicting}. Landmark-based methods fall between end-to-end and feature-based strategies, using spatial-temporal graphs of facial, body, or hand landmarks extracted through tools such as OpenFace or MediaPipe \cite{lugaresi2019mediapipe}. These graphs are processed by Spatial-Temporal Graph Convolutional Networks (ST-GCNs) \cite{yan2018spatial}, which capture dynamic patterns in engagement. ST-GCNs, initially used in facial affect analysis \cite{liu2022graph}, have been effectively applied to engagement estimation; for instance, Abedi and Khan \cite{abedi2024engagement} used MediaPipe to extract facial landmarks from the EngageNet dataset \cite{engagenet} and applied ST-GCNs, achieving higher classification accuracy than previous approaches with significantly fewer parameters.

\subsection{Discussion}
\label{sec:related_work_discussion}
The currently available public datasets primarily focus on young student populations, often neglecting older adults and patient populations. To address this gap, we introduce the first publicly available virtual learning engagement dataset collected from older adult patients in real-world home environments. Annotated using a validated engagement annotation protocol, the dataset was created across diverse settings, e.g., varying temporal resolutions, and includes contextual information. The dataset has undergone rigorous evaluation to ensure its suitability for AI model development through machine learning and deep learning techniques. Furthermore, to maintain de-identification and adhere to findings highlighting the superiority of landmarks-based algorithms, such as ST-GCN, over traditional video-based end-to-end models, the dataset includes facial, hand, and body landmarks, along with features derived from these landmarks.

\section{OPEN: Older adult Patient ENgagement Dataset}
\label{sec:dataset}
\subsection{Setting: Virtual Rehabilitation}
\label{sec:virtual_cardiac_rehabilitation}
Following discharge from acute care after a cardiac event or procedure, eligible patients are often referred to outpatient cardiac rehabilitation programs \cite{association2009canadian}. These programs involve interdisciplinary, comprehensive risk-reduction interventions aimed at improving the physical, psychological, and social functioning of individuals with cardiovascular disease. Completion of cardiac rehabilitation has been shown to significantly reduce morbidity, as well as cardiac-specific and all-cause mortality \cite{mcdonagh2023home}. Outpatient cardiac rehabilitation can be delivered virtually, in hybrid formats, or in person, depending on patient preference \cite{ganeshan2022clinical}. However, during the final months of the COVID-19 pandemic, virtual delivery was the only available option at the time OPEN was collected.

In virtual cardiac rehabilitation, educational sessions are a key component during which OPEN was collected. These sessions typically involve groups of patients participating virtually with a clinician who provides guidance on self-management and the adoption of a heart-healthy lifestyle. Topics commonly addressed include understanding one’s diagnosis, medication adherence, balanced nutrition, physical activity, avoidance of harmful behaviors such as smoking, reduction of sedentary behavior, stress management, and strategies for incorporating regular exercise into daily routines. The primary goal is to equip patients with the knowledge and skills necessary to establish sustainable habits that support long-term health and well-being \cite{cardiaccollege}.

\begin{figure}[ht]
    \centering
    \includegraphics[width=.48\textwidth]{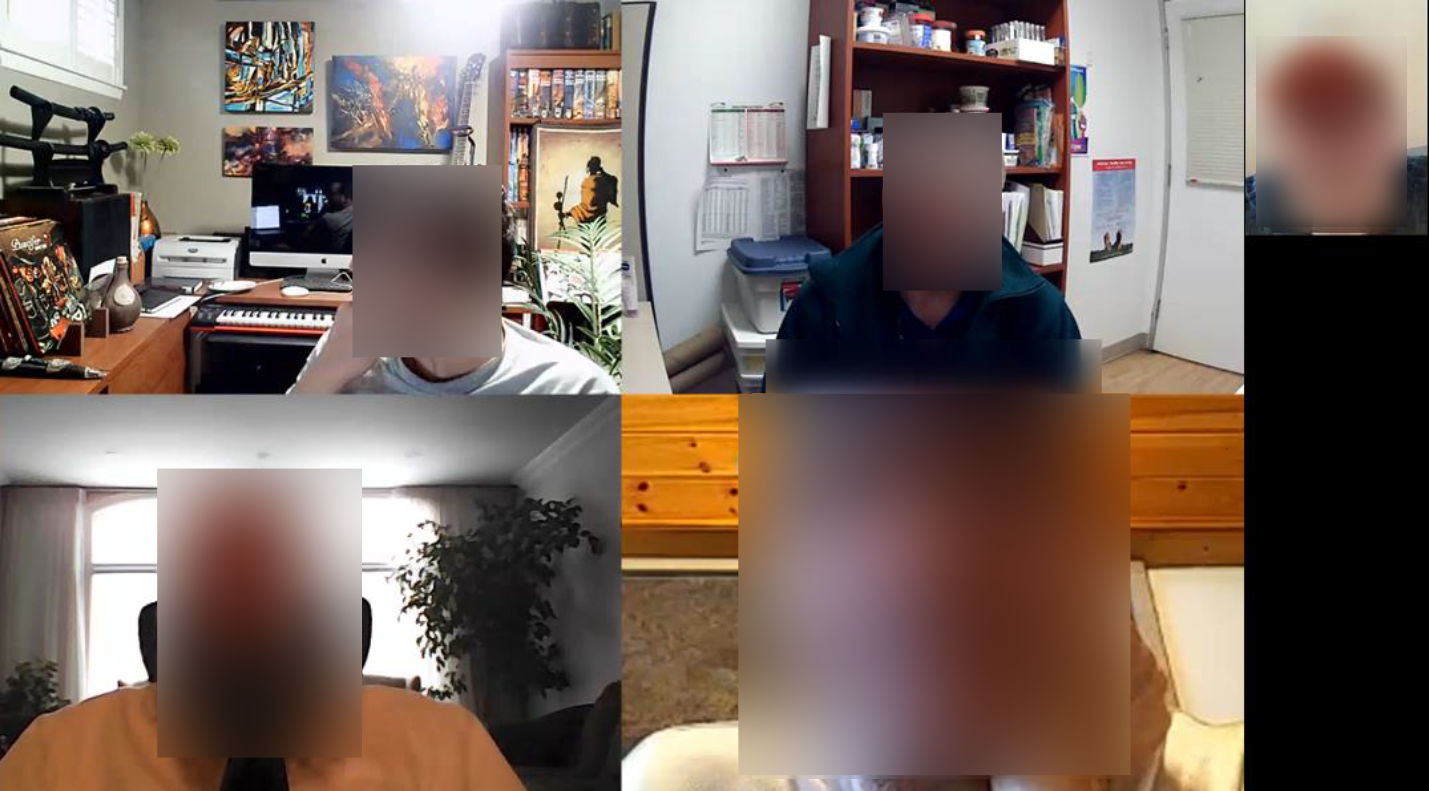}
    \caption{Screenshot of a virtual cardiac rehabilitation educational session conducted via Microsoft Teams. The session is facilitated by a cardiac rehabilitation clinician (second from the left in the top row), with four participating patients. Faces are blurred to protect participant privacy.}
    \label{fig:session}
\end{figure}

The educational sessions of virtual cardiac rehabilitation spanned six weeks, with weekly one-hour sessions scheduled on a predetermined weekday and facilitated by a cardiac rehabilitation supervisor or clinician. Groups of four to eight participants attended these sessions virtually from their homes using Microsoft Teams. During the sessions, participants remained seated and viewed the content on their personal computer or laptop screens (see Figure~\ref{fig:session}); however, they were not always continuously attentive, as moments of disengagement could occur. The sessions were primarily structured around the clinician presenting information, often using educational slides and addressing the group as a whole. The clinician also facilitated interaction by posing questions to the group or to individual participants.

Participants viewed the live video of the clinician or their shared screen displaying educational slides, both of which were highlighted using Microsoft Teams’ spotlight feature. This functionality pins the selected video or content to the main screen for all attendees, ensuring that the clinician or their material remained prominently visible throughout the session. In addition, participants could see live video feeds of themselves and other attendees, captured via their devices' built-in RGB cameras or external webcams. Audio communication was facilitated through the devices' integrated speakers and microphones, allowing participants to hear the clinician and fellow attendees and to speak when appropriate. None of the participants used earphones or headsets during the sessions.

\subsection{Participants}
\label{sec:participants}
Participants were individuals diagnosed with coronary artery disease, a cardiac event, or a cardiac procedure, primarily older adults aged 60 years and above; however, a few participants were under the age of 60. Eligibility required enrollment in a home-based virtual cardiac rehabilitation program delivered through the Cardiovascular Prevention and Rehabilitation Program at the Toronto Rehabilitation Institute, University Health Network (Toronto, Canada). Participants also needed access to a personal computer or tablet with an Internet connection. Exclusion criteria included post-procedure complications (e.g., stroke), psychiatric or cognitive impairments that could hinder participation (e.g., difficulty understanding or following instructions), and lack of fluency in English.

\subsection{Ethics Approval and Consent}
\label{sec:ethics}
The study protocol for dataset collection, annotation, predictive modeling, and public dataset release was approved by the University Health Network Research Ethics Board (Study ID: 21-5420). Informed consent was obtained from all participants whose data were included in the dataset.

\subsection{Data Collection}
\label{sec:data_collection}
The data collection process did not alter any aspect of the virtual cardiac rehabilitation educational sessions, which followed the standard of care, apart from the recording of participant videos. At the beginning of each weekly session, the clinician reminded participants to keep their cameras on and informed them that the session would be recorded via Microsoft Teams.

The recorded videos had a resolution of 1280×720 pixels and a frame rate of 16 frames per second (fps). Microsoft Teams does not support recording individual participant video streams separately; therefore, each session was captured as a single composite video file. Some participants in these sessions did not consent to study participation. Although their video data was initially recorded, it was subsequently excluded from the dataset.

\subsection{Data Annotation}
\label{sec:data_annotation}
Data annotation consisted of three tasks: spatial-temporal annotation of participants' tiles, primary engagement annotation, and temporal annotation of context types, as outlined below.

\subsubsection{Spatial-temporal Annotation of Participants' Tiles}
\label{sec:tile_annotation}
As detailed in subsection \ref{sec:data_collection}, using the Microsoft Teams video recording feature, the video of all participants in a session was recorded as a single video file, displaying separate tiles for each participant. Two annotators examined the recorded session videos to annotate the coordinates (in pixels) of participants' tiles throughout the session whenever their positions changed. Notably, the coordinates of tiles could shift during sessions, such as when the clinician began sharing their screen or when a participant temporarily turned off their camera. The first annotator conducted the initial annotations, and the second annotator reviewed and refined them. The timings and coordinates of participant tiles for each session were used to generate a single video file for each individual participant per session, containing only their respective video content.

\subsubsection{Engagement Annotation}
\label{sec:engagement_annotation}
The Human Expert Labeling Process (HELP) \cite{aslan2017human}, an engagement annotation protocol tailored for learning environments, was employed to annotate engagement in OPEN. This protocol was chosen for its alignment with the definition of engagement in educational psychology \cite{khan2022inconsistencies,khan2024revisiting}. HELP was structured into three stages: pre-annotation, annotation, and post-annotation.

In the pre-annotation stage, training was conducted for three annotators. They were provided with a chunk of the dataset, consisting of one virtual cardiac rehabilitation session, for practice annotation. Their work was reviewed, feedback was given, ambiguities were resolved, and their questions were addressed to ensure clarity.

During the annotation stage, the three annotators, referred to as \textit{sources} (see subsection \ref{sec:engagement_annotation_dimensions}), independently annotated the dataset by retrospectively watching recorded videos, representing the \textit{timing} and \textit{data modality} dimensions of engagement annotation. The videos were the sole \textit{data modality}, with no accompanying audio or contextual information.

Most previous datasets segmented participants' recorded videos into fixed-length intervals before annotation \cite{khan2022inconsistencies}, such as 10-second segments in DAiSEE \cite{gupta2016daisee} and EngageNet \cite{engagenet}. Since engagement annotation occurs after segmentation, segments can contain multiple engagement states; for instance, a participant might be engaged at the start but disengaged by the end of the segment. This overlap can create challenges for annotators and confusion for AI models trained on such data. To prevent this confusion, following HELP, OPEN employed adaptive \textit{temporal resolution} for annotation, where data segmentation was guided by annotation. Annotators watched the videos and marked the precise moments (accurate to seconds) when participants' states changed. This approach ensured that each data segment represented a single, consistent state.

Regarding the \textit{level of abstraction}, engagement was annotated as a multi-component variable. Since video was the sole data modality available to annotators, they could annotate the affective and behavioral components of engagement but not cognitive engagement~\cite{booth2023engagement}. Drawing inspiration from Woolf et al. \cite{woolf2009affect}, HELP identified the most optimal classes for engagement component annotation. These included four affective (emotional) components: Bored, Calm/Satisfied, Confused/Frustrated, and Motivated/Excited,  as well as two behavioral components: Off-Task and On-Task. These classes were adopted for annotating OPEN.

Following the framework proposed by Woolf et al. \cite{woolf2009affect}, HELP established rules for the \textit{quantification} of engagement as a dichotomous variable: Engaged or Not-Engaged, based on the \textit{combination} of affective and behavioral components. If the behavioral component is Off-Task, the state is quantified as Not-Engaged. When the behavioral component is On-Task, the state is also quantified as Not-Engaged if the affective component is Bored. In all other combinations, the state is quantified as Engaged.

\subsubsection{Temporal Annotation of Context Types}
\label{sec:context_type_annotation}
Two annotators reviewed the recorded session videos to identify the start and end times of various context types. The first annotator conducted the initial annotations, and the second annotator reviewed and refined them. These context types included: (i) the clinician addressing all participants, (ii) the clinician speaking to a specific participant (with the participant identified), (iii) a participant speaking to the clinician, (iv) the clinician presenting slides or videos, (v) a participant addressing all other participants, (vi) a participant speaking to another participant, and (vii) periods of inactivity or absence of interaction.

\subsection{Dataset Characteristics}
\label{sec:dataset_characteristics}
\subsubsection{Participant Characteristics}
\label{sec:participant_characteristics}
Table~\ref{tab:participant_demographics} presents the demographic characteristics of the 11 participants included in OPEN. The participants had a mean age of 66.5 years (SD = 9.6). Of the total, 36.4\% (n = 4) were female, 90.9\% (n = 10) identified as Caucasian, and 9.1\% (n = 1) identified as South Asian. Participants experienced a diverse range of cardiac events and procedures.

\begin{figure}[ht]
    \centering
    \includegraphics[width=.5\textwidth]{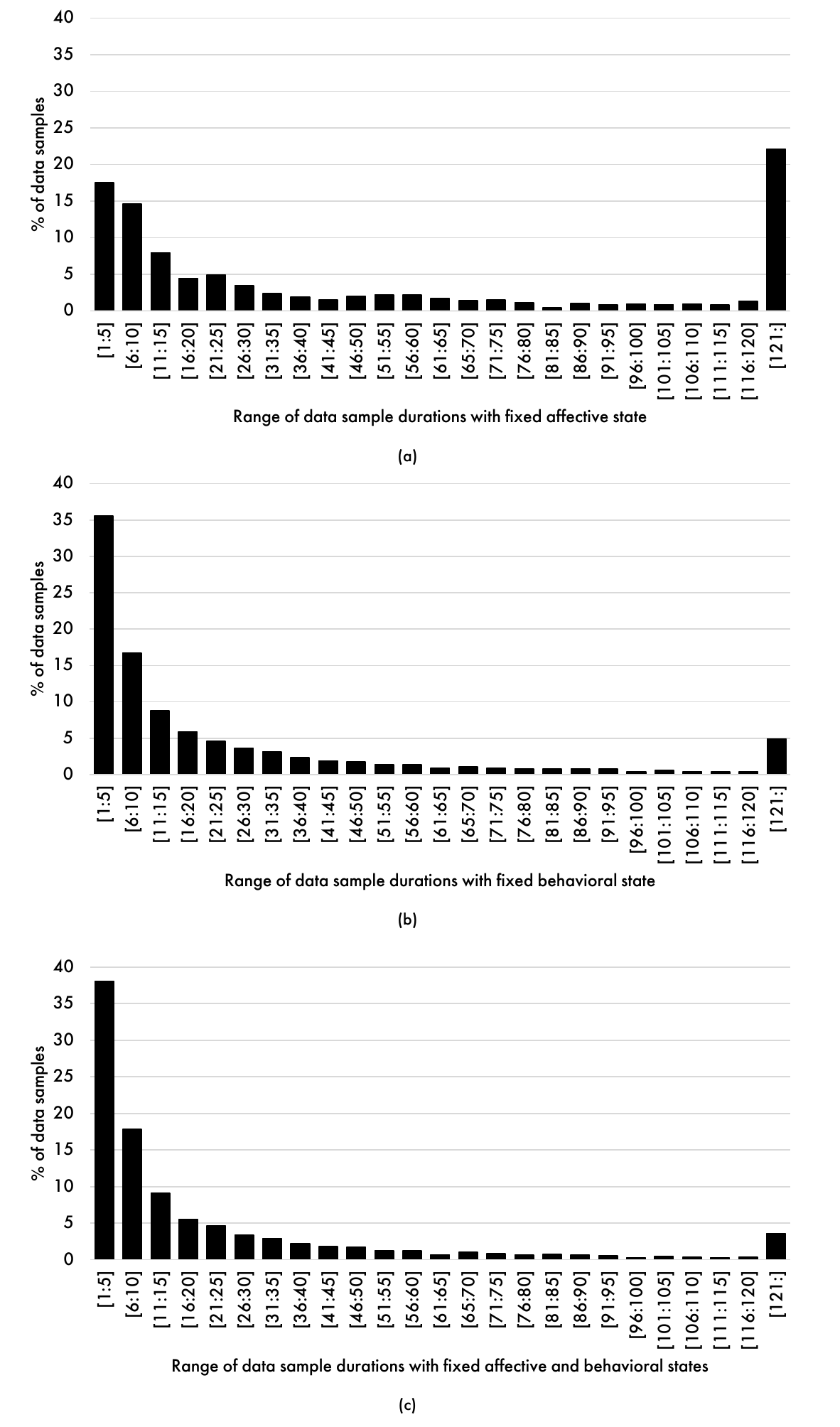}
    \caption{The frequency of data samples across various ranges of sample durations (in seconds) with fixed (a) affective, (b) behavioral, and (c) both affective and behavioral states. For instance, the first data point shows the duration range of 1 to 5 seconds. The last data point shows the duration range of longer than 2 minutes.}
    \label{fig:percentages}
\end{figure}

\begin{table*}[t]
\caption{Participant demographics, cardiac diseases/procedures, number of virtual sessions (and the amount of data) in the dataset.}
\centering
\begin{tabular}{lp{1cm}p{1cm}p{2cm}p{5cm}p{3cm}}
\hline
\textbf{\#} & \textbf{Age} & \textbf{Sex} & \textbf{Ethnicity} & \textbf{Cardiac Disease / Procedure} & \textbf{\# of Sessions (duration)} \\ \hline
1  & 62 & Male   & Caucasian   & Aortic Valve Replacement & 4  (03:56)\\
2  & 60 & Male   & Caucasian   & Coronary Artery Bypass Graft & 6 (05:26)\\
3  & 80 & Female & Caucasian   & Angioplasty and Stent Placement & 2 (02:14)\\
4  & 52 & Female & South Asian & Chemotherapy-induced Cardiotoxicity & 1 (00:15)\\
5  & 72 & Female & Caucasian   & Atrial Fibrillation & 1 (01:08)\\
6  & 76 & Male   & Caucasian   & Transient Ischemic Attack & 5 (04:31)\\
7  & 74 & Male   & Caucasian   & Stent Placement & 5 (04:33)\\
8  & 77 & Female & Caucasian   & Aortic Valve Replacement & 3 (03:02)\\
9  & 65 & Male   & Caucasian   & Stent Placement & 2 (02:22)\\
10 & 64 & Female & Caucasian   & Stress Induced Cardiomyopathy & 4 (04:05)\\
11 & 47 & Male   & Caucasian   & Percutaneous Coronary Intervention & 3 (03:27)\\ \hline
\end{tabular}
\label{tab:participant_demographics}
\end{table*}

\subsubsection{Temporal Characteristics}
\label{sec:temporal_characteristics}
The last column of Table \ref{tab:participant_demographics} presents the number of virtual sessions attended by each participant in their virtual program, along with the total duration of their data in the dataset. The dataset comprises 36 participant-sessions, amounting to a total of 35 hours and 2 minutes of recorded data.

As described in subsection \ref{sec:engagement_annotation}, data samples were adaptively generated based on changes in the affective or behavioral components of engagement. This approach produces variable-length annotated data samples, with each sample having fixed affective and behavioral engagement components. Figures \ref{fig:percentages} (a), (b), and (c) show the percentage of data samples in the dataset across different duration ranges, with each panel respectively focusing on affective state, behavioral state, and the combination of both affective and behavioral states. As shown in Figure \ref{fig:percentages}, the most common duration range for data samples, across affective, behavioral, and combined engagement components, is 1 to 5 seconds. This indicates that, in most cases, participants' affective and behavioral states shifted every 1 to 5 seconds. The next most frequent range was 6 to 10 seconds, followed by longer durations. A comparison of Figures 1 (a) and (b) reveals that behavioral states changed more rapidly than affective states. The total number of data samples resulting from the combination of affective and behavioral states, as shown in Figure \ref{fig:percentages} (c), is 4,494. The durations of these samples have the following statistics (in seconds): a minimum of 1, a maximum of 1,133, a mean of 26.29, a standard deviation of 52.28, and a median of 9.

\subsubsection{Dynamics of Affective and Behavioral States}
\label{sec:dynamics}
Leveraging adaptive data sample creation based on changes in affective and behavioral states, OPEN is the first dataset in the literature to capture transitions between affective and behavioral states over time. Table \ref{tab:dynamics} presents the statistics on these transitions, referred to as the dynamics of affective and behavioral states in the literature \cite{d2012dynamics,d2007dynamics}. These dynamics provide valuable insights for developing models to predict future affective states, behavioral states, and levels of engagement.

To illustrate, in Table \ref{tab:dynamics}, the most frequent transitions occur between Off-Task and On-Task when the affective state is Calm/Satisfied. When participants are On-Task, the likelihood of transitioning from Calm/Satisfied to Motivated/Excited is significantly higher than transitioning to Bored or Confused/Frustrated. Additionally, transitions from Bored to Motivated/Excited while On-Task are far less likely than transitions to Calm/Satisfied. Although the dynamics of affective and behavioral states have been studied in other populations, the analysis of these dynamics in older adult patient populations is novel. Notably, the observed trends in affective state transitions during On-Task periods are consistent with prior findings in young, healthy student populations \cite{d2012dynamics}.

\begin{table*}[ht]
\caption{\justifying Dynamics of behavioral (Off-Task and On-Task) and affective (Bored, Calm/Satisfied, Confused/Frustrated, and Motivated/Excited) states. The states listed in the first column represent the current states, while the numbers in the subsequent columns indicate the frequency with which the corresponding next states occurred. For example, the state "Off-Task, Calm/Satisfied" followed the state "Off-Task, Bored" in 21 instances.}
\label{tab:dynamics}
\centering
\renewcommand{\arraystretch}{1.3} 
\resizebox{0.9\textwidth}{!}{
\begin{tabular}{lccccc|ccccc}
\toprule
& \makecell{\textbf{Off, Bored}} & \makecell{\textbf{Off,} \textbf{Calm}} & \makecell{\textbf{Off,} \textbf{Confused}} & \makecell{\textbf{Off,} \textbf{Motivated}} & & \makecell{\textbf{On,} \textbf{Bored}} & \makecell{\textbf{On,} \textbf{Calm}} & \makecell{\textbf{On,} \textbf{Confused}} & \makecell{\textbf{On,} \textbf{Motivated}} \\
\midrule
\raggedright{\textbf{Off, Bored}}               & -- & 21 & 0 & 2 & & 0 & 14 & 0 & 0 \\
\raggedright{\textbf{Off, Calm}}                & 35 & -- & 14 & 87 & & 1 & 1586 & 0 & 22 \\
\raggedright{\textbf{Off, Confused}}            & 1 & 16 & -- & 1 & & 0 & 1 & 6 & 0 \\
\raggedright{\textbf{Off, Motivated}}           & 2 & 101 & 0 & -- & & 0 & 20 & 0 & 76 \\
\midrule 
\raggedright{\textbf{On, Bored}}                & 18 & 0 & 0 & 0 & & -- & 17 & 0 & 3 \\
\raggedright{\textbf{On, Calm}}                 & 13 & 1576 & 2 & 27 & & 3 & -- & 5 & 232 \\
\raggedright{\textbf{On, Confused}}             & 0 & 0 & 8 & 0 & & 0 & 4 & -- & 0 \\
\raggedright\textbf{On, Motivated}           & 1 & 24 & 0 & 92 & & 1 & 208 & 0 & -- \\
\bottomrule
\end{tabular}%
}
\end{table*}

\begin{table*}[t]
\caption{Distribution of data samples across behavioral, affective, and engagement states in different settings.}
\centering
\begin{tabular}{lc|ccc|ccccccc}
\hline
\textbf{Class} & \makecell{\textbf{Variable-}\\\textbf{length}} & \makecell{\textbf{5 seconds-}\\\textbf{strategy 1}} & \makecell{\textbf{10 seconds-}\\\textbf{strategy 1}} & \makecell{\textbf{30 seconds-}\\\textbf{strategy 1}} & \makecell{\textbf{5 seconds-}\\\textbf{strategy 2}} & \makecell{\textbf{10 seconds-}\\\textbf{strategy 2}} & \makecell{\textbf{30 seconds-}\\\textbf{strategy 2}} \\
\midrule
\textbf{Off-Task}           & 2144 & 9544  & 4621  & 1580  & 10787 & 5747  & 2543  \\
\textbf{On-Task}            & 2350 & 12886 & 6335  & 2207  & 11643 & 5209  & 1244  \\
\midrule
\textbf{Bored}              & 113  & 511   & 248   & 80    & --    & --    & --    \\
\textbf{Calm/Satisfied}      & 3822 & 20952 & 10295 & 3601  & --    & --    & --    \\
\textbf{Confused/Frustrated} & 38   & 74    & 38    & 11    & --    & --    & --    \\
\textbf{Motivated/Excited}   & 521  & 893   & 375   & 95    & --    & --    & --    \\
\midrule
\textbf{Engaged}            & 2214 & 12828 & 6308  & 2201  & 11603 & 5194  & 1243  \\
\textbf{Not-Engaged}        & 2280 & 9602  & 4648  & 1586  & 10827 & 5762  & 2544  \\
\midrule
\textbf{Total}              & 4494 & 22430 & 10956 & 3787  & --    & --    & --    \\
\midrule
\end{tabular}
\label{tab:class_distribution}
\end{table*}

\subsubsection{Fixed-length Data Sample Creation}
\label{sec:data_sample_creation}
In addition to the variable-length data samples in the dataset, two methods were implemented to generate fixed-length data samples. After segmenting the dataset into fixed-length data samples of length $l$ seconds, if $l$ spans more than one second, some data samples may overlap with transitions between different participant affective and behavioral states. Two strategies were employed to determine the participant state, or label, assigned to each fixed-length data sample. The first strategy for labeling data segments of size $l$ involved majority voting. In this approach, the affective and behavioral states with the highest occurrence within a data sample were selected as the corresponding labels for that data sample. The second strategy leveraged the principle that if a participant's behavioral state is Off-Task for any moment within a data sample, the entire data sample should be labeled as Off-Task for the behavioral state. The affective state, however, was still determined using majority voting.

\subsubsection{Class Distributions}
\label{sec:class_distributions}
Table~\ref{tab:class_distribution} presents the distribution of data samples across behavioral, affective, and engagement states for various settings: original variable-length samples and fixed-length samples of 5, 10, and 30 seconds, created using Strategies 1 and 2 as described in Subsection~\ref{sec:data_sample_creation}. In the variable-length, 5-second, and 10-second settings, the distributions across affective states are imbalanced, with a strong skew toward Calm/Satisfied. However, the distributions across behavioral and engagement states are relatively balanced in these settings. In contrast, for the 30-second settings, the distributions across behavioral and engagement states become imbalanced. When using Strategy 2 compared to Strategy 1, there is an increase in Off-Task samples, which subsequently results in more Not-Engaged data samples. The total number of data samples is 4,494 in the variable-length setting and 22,430, 10,956, and 3,787 in the 5-, 10-, and 30-second fixed-length settings, respectively.

There are both similarities and differences between the OPEN dataset and the existing public engagement datasets, such as DAiSEE \cite{gupta2016daisee} and EngageNet \cite{engagenet}, in terms of the distribution of data samples across affective and behavioral components of engagement, as well as engagement levels. The OPEN dataset was collected from older adult patients at home over consecutive weekly sessions, whereas DAiSEE and EngageNet were collected from young students in controlled settings during a single 20 to 30-minute session. In DAiSEE, engagement was annotated solely as an affective state, disregarding behavioral aspects. In contrast, EngageNet considered both affective and behavioral components in engagement annotations.

A key similarity between DAiSEE with 10-second data samples and OPEN in the 10-second length data sample configuration lies in the imbalance of affective engagement annotations. In DAiSEE, 95.6\% of the samples are labeled as high engagement. Similarly, in OPEN, the affective component is heavily skewed toward the Calm/Satisfied state (93.96\% of data samples), which is typically associated with being Engaged \cite{woolf2009affect,aslan2017human}. Another point of similarity can be observed in the overall engagement distribution between EngageNet and OPEN. EngageNet, which annotated 10-second data samples based on both affective and behavioral cues, does not display a strong class imbalance, 30.5\% low engagement and 69.5\% high engagement. This is comparable to the overall engagement distribution in OPEN with 10-send data samples (42.4\% Not-Engaged versus 57.6\% Engaged), which also incorporates both affective and behavioral components following the HELP protocol. In other configurations of OPEN, outlined in Table \ref{tab:class_distribution}, such as when using 30-second data samples, the distribution of engagement states differs from both DAiSEE and EngageNet.

\subsubsection{Inter Rater Reliability}
\label{sec:inter_rater_reliability}
As outlined in subsection \ref{sec:engagement_annotation}, the annotation procedure involved creating new data samples based on changes in the affective or behavioral components of engagement, accurate to the second. Consequently, every second of the dataset includes annotations from three annotators.

Fleiss' Kappa of 0.8254, Krippendorff's Alpha of 0.8434, and pairwise Cohen's Kappa of 0.7376, 0.8492, and 0.8891 were achieved for behavioral engagement annotation, which included three states: Off-Task, On-Task, and Not-Visible.

For affective engagement annotation, involving five states (Bored, Calm/Satisfied, Confused/Frustrated, Motivated/Excited, and Not-Visible), Fleiss' Kappa was 0.7124, Krippendorff's Alpha was 0.5746, and pairwise Cohen's Kappa values were 0.6662, 0.7695, and 0.7014.

Engagement annotation with three states (Engaged, Not-Engaged, and Not-Visible) resulted in Fleiss' Kappa of 0.8108, Krippendorff's Alpha of 0.8107, and pairwise Cohen's Kappa of 0.7172, 0.8448, and 0.8704 \cite{gwet2014handbook}.

In most cases, agreement values ranged between 0.61–0.80 (substantial agreement) and 0.81–1.00 (almost perfect agreement) \cite{mchugh2012interrater}, surpassing the levels reported in the original HELP \cite{aslan2017human} and prior datasets \cite{engagenet}. Consistent with trends in HELP \cite{aslan2017human}, affective engagement annotations exhibited lower agreement compared to behavioral engagement annotations. This is attributed to the inherently subjective nature of affective engagement annotation \cite{aslan2017human} and the greater number of possible states.

\subsubsection{Context Type Distribution}
\label{sec:context_type_distribution}
\begin{figure}[ht]
    \centering
    \includegraphics[width=.5\textwidth]{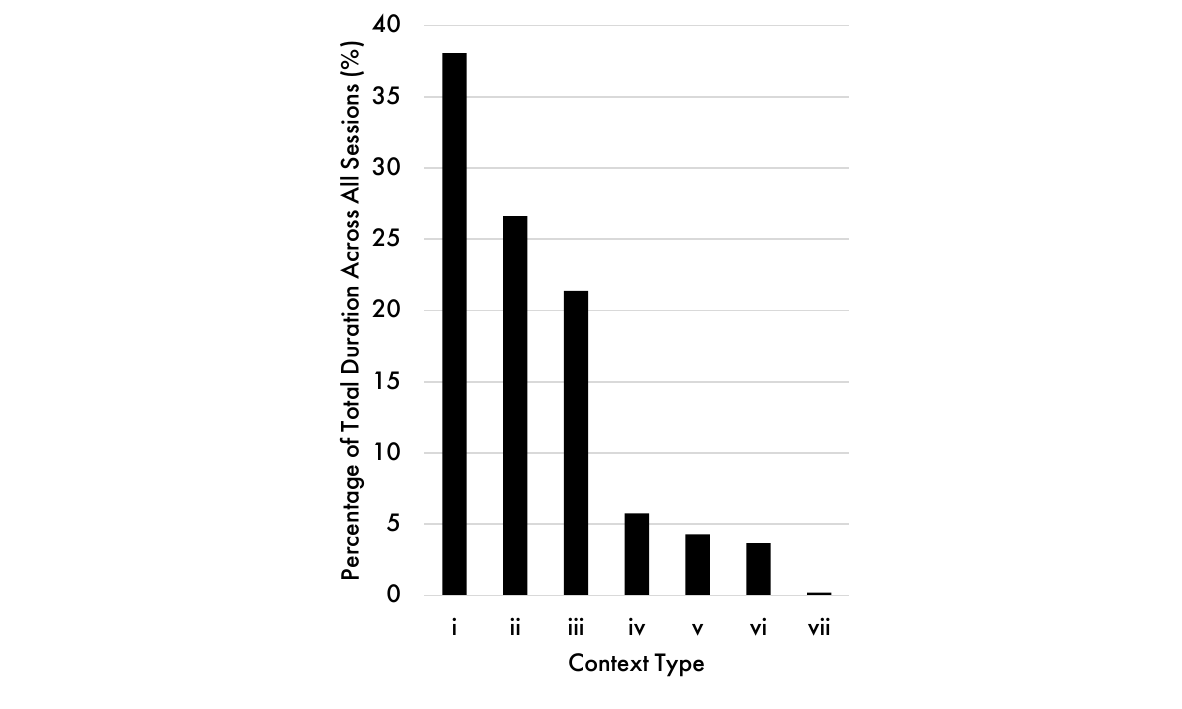}
    \caption{Distribution of the durations of the seven context types, as described in subsection \ref{sec:context_type_annotation}, aggregated across all sessions.}
    \label{fig:context_type_distribution}
\end{figure}

Figure \ref{fig:context_type_distribution} illustrates the distribution of the seven context types described in subsection \ref{sec:context_type_annotation}. Across all sessions, the majority of time (64.69\%) is devoted to the first two context types, when the clinician is speaking to all participants or to a specific individual. In 21.39\% of the time, a participant is speaking to the clinician, while the remaining 13.92\% corresponds to all other context types. Notably, each participant (numbered 1 to 11) is identified in the annotations, enabling analysis of how different context types influence individual participant engagement throughout the sessions.

\subsection{Features in the Dataset}
\label{sec:dataset_features}
As previously noted, the raw audio-video data was withheld to protect patient privacy; however, facial, hand, and body joint landmarks, along with behavioral and affective features derived from these landmarks, were extracted from the video. The following features were obtained from each video frame.

Using OpenFace \cite{baltrusaitis2018openface}, 668 features were extracted, including: six for eye gaze direction vectors in world coordinates, two for gaze direction in radians, 112 for 2D eye landmarks (pixels), 168 for 3D eye landmarks (millimeters), three each for head location and rotation, 68 x 2 for 2D facial landmarks (pixels), 68 x 3 for 3D facial landmarks (millimeters), and 34 for the occurrence and intensity of 17 FAUs numbered 1, 2, 4, 5, 6, 7, 9, 10, 12, 14, 15, 17, 20, 23, 25, 26, and 45.

Using the pre-trained EmoFAN \cite{toisoul2021estimation}, 2 values of valence and arousal were extracted from facial regions of each video frame.

The MediaPipe deep learning framework \cite{lugaresi2019mediapipe}, known for its real-time and cross-platform capabilities, was utilized to extract facial, hand, and body landmarks from video. Within MediaPipe, the Attention Mesh model \cite{grishchenko2020attention} accurately detects 468 3D facial landmarks across the face, along with an additional 10 landmarks for the iris, resulting in a total of \(478 \times 3\) 3D landmarks for the face and iris. MediaPipe also provides \(21 \times 3\) 3D landmarks for hand tracking and \(33 \times 3\) 3D landmarks for body joints.

The format of a data sample in the dataset is a CSV file where each row corresponds to a frame of the data sample, and the columns represent the aforementioned features extracted from the frame. These are followed by the ground-truth affective and behavioral components of engagement, as well as the engagement annotation associated with the frame.

\section{Technical Validation: Engagement Recognition}
\label{sec:experiments}
For benchmarking and technical validation of OPEN, a range of machine learning and deep learning models, recognized for their superior performance on previous datasets in the literature \cite{karimah2022automatic,abedi2024engagement,vedernikov2024tcct,engagenet}, were developed. Given the diverse data types and varying data sample lengths in OPEN, these models were trained and evaluated to ensure alignment with the specific characteristics of the respective input data types.

In alignment with the engagement annotation in OPEN, engagement inference is formulated as a binary classification task. For the first time in the field of engagement recognition \cite{karimah2022automatic,abedi2024engagement,vedernikov2024tcct,engagenet}, two distinct types of engagement inference tasks were systematically investigated: \textit{engagement detection}, a task extensively addressed in existing literature, which focuses on estimating engagement at the current timestamp using data from the same timestamp, and \textit{engagement prediction}, a novel task that involves estimating engagement at a future timestamp based on data from a preceding timestamp.

Furthermore, for the first instance in the engagement recognition domain \cite{karimah2022automatic,abedi2024engagement,vedernikov2024tcct,engagenet}, the performance of various models and feature sets was systematically evaluated across different dataset settings, considering data sample lengths of 5 seconds, 10 seconds, 30 seconds, and variable lengths, as described in subsection \ref{sec:class_distributions}.

Additionally, for the first time in the field of engagement recognition \cite{karimah2022automatic,abedi2024engagement,vedernikov2024tcct,engagenet}, two distinct approaches to engagement estimation were compared: developing models that directly output engagement as a dichotomous variable, which has been the focus of existing literature, and a novel approach involving the development of models that output the affective and behavioral components of engagement, from which engagement can be subsequently derived.

A few subsets of the dataset features, outlined in subsection \ref{sec:dataset_features}, were utilized for engagement recognition in line with the existing literature \cite{engagenet,vedernikov2024tcct,abedi2024engagement}. Different combinations of the following features, extracted from consecutive video frames, were used as input to machine-learning and deep-learning models capable of analyzing sequential data. These features include 6 eye gaze features, 6 head pose (location and rotation) features, the intensities of 17 FAUs \cite{engagenet}, and 2 values of valence and arousal \cite{abedi2023affect} as affect features. In addition to the aforementioned features, facial landmarks, as detailed in subsection \ref{sec:dataset_features}, were also used for engagement estimation. The models included LSTM, ST-GCN \cite{yan2018spatial}, Transformer, and ROCKET \cite{dempster2020ROCKET}.

The LSTM model consists of a two-layer LSTM architecture, each layer containing 64 hidden units. This is followed by a fully connected layer of size 64 × 1, where the output dimension of 1 corresponds to the binary classification task. The Transformer model comprises two layers of Transformer encoders with an input dimension of 64, four attention heads, and 128 as the dimension of the feedforward network model. The input features are first processed through a linear projection layer that maps the original input dimension to 64 before being passed to the Transformer encoder. The ST-GCN consists of three ST-GCN layers, followed by an average pooling layer. The output from the average pooling layer is further processed by a convolutional layer that projects it to a single output dimension for classification. All deep learning models were optimized using the Adam optimizer for 1000 epochs with an initial learning rate of 0.0001, which was reduced by a factor of 10 every 250 epochs. The ROCKET model employs 10,000 randomly initialized convolutional kernels to extract feature representations. The transformed feature space generated by ROCKET is subsequently classified using a Categorical Boosting (CatBoost) model. The CatBoost classifier was trained for 1000 iterations with a maximum tree depth of 6.

The evaluation metrics for engagement inference, framed as a binary classification problem, include accuracy, precision, recall, F1-score, Area Under the Curve-Receiver Operating Characteristic (AUC-ROC), and Area Under the Curve-Precision Recall (AUC-PR).

To assess the generalizability of the models on data from unseen participants, the models were trained and evaluated using two different settings of 11-fold Cross-Validation (CV) (where 11 represents the number of participants in the dataset) and Leave-One-Participant-Out (LOPO) CV. The annotation process was tailored to be as participant-independent as possible, and the data consists of features or facial landmarks extracted from video, which are less participant-dependent than raw video. However, LOPO CV is expected to yield lower performance than 11-fold CV, as the latter allows data from the same participant to appear in both the training and validation sets within a CV iteration. This is due to the fact that the manifestations of affective and behavioral indicators of engagement and disengagement are unique to each participant.

\begin{table*}[t]
\caption{Engagement detection results of different feature sets and models on OPEN. (EG: Eye Gaze, HP: Head Pose, VA: Valence-Arousal, FAU: Facial Action Units, CV: Cross-Validation, 11-fold: 11-Fold CV, LOPO: Leave-One-Participant-Out CV)}
\centering
\renewcommand{\arraystretch}{1.1}
\setlength{\tabcolsep}{6pt} 
\begin{tabular}{lllcccccc}
\toprule
\textbf{CV} & \textbf{Features} & \textbf{Model} & \textbf{Accuracy} & \textbf{Precision} & \textbf{Recall} & \textbf{F1 Score} & \textbf{AUC-ROC} & \textbf{AUC-PR} \\
\midrule
LOPO & EG, HP, VA & LSTM & 0.6437 & 0.6444 & 0.7486 & 0.6926 & 0.6633 & 0.6759 \\
LOPO & EG, HP, VA & Transformer & 0.6671 & 0.6888 & 0.6915 & 0.6902 & 0.7087 & 0.7178 \\
LOPO & EG, HP, VA & ROCKET & 0.6133 & 0.6221 & 0.7098 & 0.6631 & 0.6631 & 0.6908 \\
11-fold & EG, HP, VA & LSTM & 0.6466 & 0.6563 & 0.7155 & 0.6846 & 0.6902 & 0.7097 \\
11-fold & EG, HP, VA & Transformer & 0.7264 & 0.7731 & 0.6932 & 0.7310 & 0.7917 & 0.8120 \\
11-fold & EG, HP, VA & ROCKET & 0.7089 & 0.7260 & 0.7341 & 0.7301 & 0.7828 & 0.8011 \\
\hline
LOPO & EG, HP, VA, FAU & Transformer & 0.6086 & 0.6264 & 0.6685 & 0.6468 & 0.6199 & 0.6086 \\
LOPO & EG, HP, VA, FAU & ROCKET & 0.5925 & 0.5991 & 0.7250 & 0.6561 & 0.6313 & 0.6622 \\
11-fold & EG, HP, VA, FAU & Transformer & 0.7707 & 0.7915 & 0.7771 & 0.7842 & 0.8240 & 0.8263 \\
11-fold & EG, HP, VA, FAU & ROCKET & 0.7384 & 0.7419 & 0.7853 & 0.7630 & 0.8070 & 0.8110 \\
\hline
LOPO & Facial Landmarks & Transformer & 0.6345 & 0.6932 & 0.5707 & 0.6260 & 0.6839 & 0.7125 \\
LOPO & Facial Landmarks & ST-GCN & 0.6529 & 0.7050 & 0.5530 & 0.6198 & 0.6973 & 0.6815 \\
LOPO & Facial Landmarks & ROCKET & 0.6362 & 0.6389 & 0.7391 & 0.6854 & 0.7019 & 0.7455 \\
11-fold & Facial Landmarks & Transformer & 0.7018 & 0.7484 & 0.6685 & 0.7062 & 0.7460 & 0.7820 \\
11-fold & Facial Landmarks & ST-GCN & 0.7660 & 0.8054 & 0.7438 & 0.7734 & 0.8374 & 0.8503 \\
11-fold & Facial Landmarks & ROCKET & 0.8112 & 0.8313 & 0.8211 & 0.8261 & 0.8934 & 0.9125 \\
\bottomrule
\end{tabular}
\label{tab:engagement_detection}
\end{table*}

\subsection{Engagement Detection}
\label{sec:engagement_detection}
Table \ref{tab:engagement_detection} presents the results of engagement detection across various feature sets and models on OPEN using 10-second data samples, as described in subsection \ref{sec:class_distributions}. This duration represents the most common data sample length in the literature \cite{engagenet,gupta2016daisee,khan2022inconsistencies}. Across all feature sets and model combinations presented in Table \ref{tab:engagement_detection}, the results of 11-fold CV consistently outperform those of LOPO CV. As discussed in subsection \ref{sec:experiments}, this performance gap arises from the unique affective and behavioral states, as well as the distinct manifestations of engagement in individual participants, making it more challenging for models to generalize to unseen participants in the test sets of LOPO CV.

As shown in Table \ref{tab:engagement_detection}, when using feature sets that include eye gaze, head pose, and valence-arousal, with or without FAUs, the Transformer model achieved the best performance, followed by ROCKET and LSTM. However, when 3D facial landmarks were used for modeling, ST-GCN, specifically designed for spatio-temporal graph data, outperformed both Transformer and ROCKET in LOPO CV. Notably, in 11-fold CV, ROCKET surpassed ST-GCN, achieving the highest accuracy of 0.8112 and an AUC-ROC of 0.8934.

The original frame rate of the videos, as well as the features extracted from video frames, was 16 fps. However, the results presented in Table \ref{tab:engagement_detection} were obtained using an fps of 8, meaning every other frame was used. Table \ref{tab:engagement_detection_lengths} compares the results of 3D facial landmarks with ROCKET for 10-second data samples when the fps is 16 versus 8. As shown in Table \ref{tab:engagement_detection_lengths}, using 16 fps deteriorates performance in LOPO CV but improves results in 11-fold CV. This is because higher fps captures more nuanced details of participants' affective and behavioral states. In 11-fold CV, where data from the same participant appears in both training and test sets, these details enhance model performance. However, in LOPO CV, where participants in the test set are unseen during training, the lack of shared participant-specific details makes the additional nuances less beneficial and potentially detrimental.

The performance of the models and feature sets that achieved the highest results on OPEN with 10-second data samples in Table \ref{tab:engagement_detection_lengths} was further evaluated on OPEN with 5-second, 30-second, and variable-length data samples, as shown in Table \ref{tab:engagement_detection_lengths}. The results on OPEN with 5-second and 30-second data samples are respectively superior and inferior to those with 10-second data samples. This is primarily due to the number of available samples: OPEN with 5-second data contains twice as many samples, while OPEN with 30-second data has three times fewer compared to the 10-second version, refer to Table \ref{tab:class_distribution}. Additionally, in OPEN with 30-second data samples, the longer sequences make it more challenging for the model to effectively capture long-term dependencies. LSTM, a model capable of handling variable-length sequences, was used with eye gaze, head pose, and valence-arousal features for variable-length sequences. The variable-length sequences correspond to the original data samples annotated using the adaptive strategy, where a new sample is created whenever a change in affective or behavioral states occurs, as described in subsection \ref{sec:dynamics}. As observed, the results on variable-length samples are significantly lower than those on fixed-length samples, highlighting the need for more advanced models and alternative feature sets.

\begin{table*}[t]
\caption{Engagement detection results of different feature sets and models on OPEN with different data sample lengths and frame rates. (EG: Eye Gaze, HP: Head Pose, VA: Valence-Arousal, FAU: Facial Action Units, CV: Cross-Validation, 11-fold: 11-Fold CV, LOPO: Leave-One-Participant-Out CV,  FPS: Frames Per Second)}
\centering
\renewcommand{\arraystretch}{1.2}
\setlength{\tabcolsep}{6pt} 
\begin{tabular}{cc ll lcccccc}
\toprule
\textbf{Length} & \textbf{FPS} & \textbf{CV} & \textbf{Features} & \textbf{Model} & \textbf{Accuracy} & \textbf{Precision} & \textbf{Recall} & \textbf{F1 Score} & \textbf{AUC-ROC} & \textbf{AUC-PR} \\
\midrule
10 & 8  & LOPO   & Facial Landmarks & ROCKET & 0.6362 & 0.6389 & 0.7391 & 0.6854 & 0.7019 & 0.7455 \\
10 & 16 & LOPO   & Facial Landmarks & ROCKET & 0.6178 & 0.6248 & 0.7190 & 0.6686 & 0.6863 & 0.7370 \\
5  & 16 & LOPO   & Facial Landmarks & ROCKET & 0.6413 & 0.6604 & 0.8587 & 0.7466 & 0.6678 & 0.7768 \\
30 & 16 & LOPO   & Facial Landmarks & ROCKET & 0.5945 & 0.5608 & 0.7737 & 0.6503 & 0.7114 & 0.7419 \\
10 & 8  & 11-fold & Facial Landmarks & ROCKET & 0.8112 & 0.8313 & 0.8211 & 0.8261 & 0.8934 & 0.9125 \\
10 & 16 & 11-fold & Facial Landmarks & ROCKET & 0.8161 & 0.8361 & 0.8252 & 0.8306 & 0.8922 & 0.9147 \\
5  & 16 & 11-fold & Facial Landmarks & ROCKET & 0.8347 & 0.8384 & 0.8140 & 0.8260 & 0.9158 & 0.9131 \\
30 & 16 & 11-fold & Facial Landmarks & ROCKET & 0.7556 & 0.7830 & 0.8791 & 0.8283 & 0.8201 & 0.9023 \\
\hline
Variable & 16 & LOPO   & EG, HP, VA & LSTM & 0.5207 & 0.5166 & 0.4227 & 0.4650 & 0.5039 & 0.4940 \\
Variable & 16 & 11-fold & EG, HP, VA & LSTM & 0.5656 & 0.5577 & 0.4842 & 0.5184 & 0.5854 & 0.5614 \\
\bottomrule
\end{tabular}
\label{tab:engagement_detection_lengths}
\end{table*}

\subsection{Engagement Prediction}
\label{sec:engagement_prediction}
As explained in Section \ref{sec:dataset}, the fixed-length data samples in OPEN are created by splitting 1-hour sessions into 10-second segments. This means that within a session, a 10-second data sample can be viewed as a preceding timestamp relative to the next 10-second data sample. For engagement prediction, each data sample is assigned the engagement label of the following data sample from the same session. In this way, the model learns to predict engagement at a future timestamp based on data from the preceding one. Table \ref{tab:engagement_prediction} presents the engagement prediction results of ROCKET and Transformer using eye gaze, head pose, valence-arousal, and FAUs on OPEN with 10-second data samples. In LOPO and 11-fold CV, Transformer and ROCKET perform best, respectively, with ROCKET achieving a maximum accuracy of 0.7034.

Table \ref{tab:engagement_prediction} also includes engagement prediction results for ROCKET using 3D facial landmarks in 11-fold CV across different data sample lengths. The results on OPEN with 30-second data samples outperform 10-second samples, while 5-second data samples perform worse. This indicates that providing 3D facial landmark data over a longer period improves the model’s ability to predict engagement at future timestamps with higher accuracy.

\subsection{Engagement Component Detection}
\label{sec:engagement_components}
To assess how affective and behavioral components of engagement can be utilized for engagement detection, the best-performing model so far, ROCKET with facial landmarks, was trained on OPEN with 10-second data samples. The focus was on engagement detection in a multi-task learning setting. First, ROCKET was trained to extract feature representations from the input data without targeting any specific output. These extracted features were then used to train two separate CatBoost models for downstream classification tasks: a 4-class affective engagement component and a 2-class behavioral engagement component, as described in subsection \ref{sec:engagement_annotation}. Next, the detected affective and behavioral engagement components were combined to generate the final binary engagement label, as detailed in subsection \ref{sec:engagement_annotation}. The results for LOPO and 11-fold CV are presented in Table \ref{tab:engagement_component}. As observed, accuracy for emotional state detection was significantly higher than for behavioral state classification, and the final engagement accuracy closely aligned with that of the behavioral state classification.

\begin{table*}[t]
\caption{Engagement prediction results of different feature sets and models on OPEN with different data sample lengths. (EG: Eye Gaze, HP: Head Pose, VA: Valence-Arousal, FAU: Facial Action Units, CV: Cross-Validation, 11-fold: 11-Fold CV, LOPO: Leave-One-Participant-Out CV,  FPS: Frames Per Second)}
\centering
\renewcommand{\arraystretch}{1.2}
\setlength{\tabcolsep}{6pt} 
\begin{tabular}{c ll lcccccc}
\toprule
\textbf{Length} & \textbf{CV} & \textbf{Features} & \textbf{Model} & \textbf{Accuracy} & \textbf{Precision} & \textbf{Recall} & \textbf{F1 Score} & \textbf{AUC-ROC} & \textbf{AUC-PR} \\
\midrule
10 & LOPO   & EG, HP, VA, FAU & Transformer & 0.5908 & 0.6128 & 0.6533 & 0.6324 & 0.5922 & 0.6241 \\
10 & LOPO   & EG, HP, VA, FAU & Rocket      & 0.5686 & 0.5766 & 0.7346 & 0.6461 & 0.5891 & 0.6143 \\
10 & 11-fold & EG, HP, VA, FAU & Transformer & 0.6854 & 0.7117 & 0.6943 & 0.7029 & 0.7347 & 0.7478 \\
10 & 11-fold & EG, HP, VA, FAU & Rocket      & 0.7034 & 0.7070 & 0.7629 & 0.7339 & 0.7646 & 0.7791 \\
\hline
10 & 11-fold & Facial Landmarks & Rocket     & 0.6102 & 0.6157 & 0.7258 & 0.6662 & 0.6565 & 0.6693 \\
5  & 11-fold & Facial Landmarks & Rocket     & 0.5924 & 0.5687 & 0.6409 & 0.6029 & 0.6478 & 0.6493 \\
30 & 11-fold & Facial Landmarks & Rocket     & 0.6457 & 0.6891 & 0.8655 & 0.7673 & 0.5933 & 0.7557 \\
\bottomrule
\end{tabular}
\label{tab:engagement_prediction}
\end{table*}

\begin{table}[t]
\caption{Detection of behavioral and emotional components of engagement, as well as overall engagement detection on OPEN, using ROCKET with 3D facial landmarks trained in the multi-task learning setting described in subsection \ref{sec:engagement_components}. (LOPO: Leave-One-Participant-Out cross-validation, 11-fold: 11-fold cross-validation)}
\centering
\renewcommand{\arraystretch}{1.2}
\setlength{\tabcolsep}{6pt} 
\begin{tabular}{lcc}
\toprule
\textbf{Target – Metric} & \textbf{LOPO} & \textbf{11-fold} \\
\midrule
Behavioral – Accuracy   & 0.6052  & 0.8180  \\
Behavioral – Precision  & 0.5339  & 0.8046  \\
Behavioral – Recall     & 0.8175  & 0.8040  \\
Behavioral – F1 Score   & 0.6459  & 0.8043  \\
Behavioral – AUC-ROC    & 0.6491  & 0.8946  \\
Behavioral – AUC-PR     & 0.5168  & 0.8729  \\
Emotional – Accuracy    & 0.9165  & 0.9462  \\
Engagement – Accuracy   & 0.6059  & 0.8210  \\
\bottomrule
\end{tabular}
\label{tab:engagement_component}
\end{table}

\section{Conclusion and Future Work}
\label{sec:conclusion}
This paper introduced a unique dataset for developing machine learning and deep learning models for engagement estimation. The dataset is distinct in several ways. It is collected from older adult patients and includes both sequential engagement data from individual sessions and longitudinal engagement data spanning multiple sessions. It captures both session-level contextual information and participant-level data. Additionally, it contains both affective and behavioral states alongside engagement, with annotations provided by observers. The data samples feature adaptive timestamps, triggered by shifts in a person’s affective or behavioral states, enabling both engagement detection and engagement prediction. The dataset is also the largest publicly available engagement dataset and ensures privacy protection by including only non-identifiable facial, hand, and body joint landmarks, along with behavioral and affective features derived from these landmarks. For initial technical validation, various models were trained on different features available in the dataset, yielding promising results with engagement detection and prediction accuracies as high as 81\% and 70\%, respectively. However, the dataset has certain limitations, including a relatively small number of participants, low demographic diversity, and the absence of audio and raw video data. Additionally, in the technical validation section of the paper, not all unique features of the dataset were fully explored.

For future work, session-level data and context type information will be leveraged for context-oriented and participant-in-context engagement estimation. Longitudinal engagement data across multiple sessions of the virtual educational program, as well as engagement within individual sessions, will be analyzed for both longitudinal and session-level engagement assessment. Moreover, advanced models, particularly those capable of handling variable-length sequences, will be explored for engagement detection and prediction. Advanced multi-task learning and multi-label learning techniques will be employed to estimate engagement from its components. Strategies will also be developed to address the imbalanced distribution of samples in the affective component of engagement. Approaches for developing more generalizable models that achieve adequate performance on unseen participants will be investigated. Multimodal large language models will be explored for automatic analysis of landmark data, inference of emotional and behavioral states, and integration of learning session contextual information, with the goal of enhancing the performance and interpretability of the engagement recognition pipeline. Finally, the applicability of the developed models across different datasets will be investigated, particularly since previous datasets were primarily collected from young, healthy students, whereas the dataset introduced in this paper focuses on older adult patients.

\ifCLASSOPTIONcompsoc
  \section*{Acknowledgments}
\else
  \section*{Acknowledgment}
\fi
This research is funded through the New Frontiers Research Fund and J.P. Bickell Foundation Medical Research grants. The authors extend their sincere thanks and appreciation to Christopher Ho, Azra Ince, Aliana Jamal, and Ahmed Mokhtar for their invaluable contributions to the observational annotation of the dataset.

\section*{Data Availability}
Researchers interested in obtaining the OPEN dataset introduced in this paper for research purposes should contact the principal investigator at shehroz.khan@uhn.ca. Access to the dataset will be granted following the execution of a data-sharing agreement that ensures compliance with ethical standards.

\section*{Author contributions statement}
S.S.K. and T.J.F.C. conceived the study. A.A., with support from S.S.K. and T.J.F.C., prepared the research ethics board documents for study approval and dataset release.
S.S., with support from A.A., S.S.K., and T.J.F.C., obtained participant consent and collected the data. A.A. designed the data annotation protocol, while S.S., with A.A., S.S.K., and T.J.F.C.'s support, managed the data annotation process.
A.A. preprocessed, curated, and created the dataset in various settings as described in the manuscript. A.A. conducted all machine learning and deep learning experiments for dataset technical validation. A.A. drafted all sections of the manuscript. All co-authors reviewed the manuscript.

\ifCLASSOPTIONcaptionsoff
  \newpage
\fi

\bibliographystyle{IEEEtran}
\bibliography{references}


\end{document}